# Epsilon-Safe Planning

**Robert P. Goldman**
Honeywell Technology Center
MN 65-2200
3660 Technology Drive
Minneapolis, MN 55418
goldman@src.honeywell.com

**Mark S. Boddy**
Honeywell Technology Center
MN 65-2200
3660 Technology Drive
Minneapolis, MN 55418
boddy@src.honeywell.com

## Abstract

We introduce an approach to high-level conditional planning we call $\epsilon$-*safe planning*. This probabilistic approach commits us to planning to meet some specified goal with a probability of success of at least $1 - \epsilon$ for some user-supplied $\epsilon$. We describe several algorithms for $\epsilon$-safe planning based on conditional planners. The two conditional planners we discuss are Peot and Smith's nonlinear conditional planner, CNLP, and our own linear conditional planner, PLINTH. We present a straightforward extension to conditional planners for which computing the necessary probabilities is simple, employing a commonly-made but perhaps overly-strong independence assumption. We also discuss a second approach to $\epsilon$-safe planning which relaxes this independence assumption, involving the incremental construction of a probability dependence model in conjunction with the construction of the plan graph.

## 1 Introduction

In order to apply planning methods to real-world problems, we must address the problem of planning under uncertainty. There have been two previous approaches to this problem. The first has been to extend classical linear and nonlinear planners to conditional planners. Such conditional planners treat uncertainty as disjunction; the planner is assumed to be able to specify alternate outcomes for a given action (sensing is treated as an action), but not to have any information about the relative likelihood of these outcomes. CNLP [Peot and Smith, 1992], PLINTH [Goldman and Boddy, 1994a, Goldman and Boddy, 1994b] and Cassandra [Pryor and Collins, 1993] are disjunctive planners of this type. The SENSp planner extends this approach by distinguishing between sensation actions and actions which alter the world, essentially imposing a qualitative cost measure over conditional plans [Etzioni *et al.*, 1992]. The second approach, which has not often been implemented for complicated domains, is fully decision-theoretic planning. This requires a model which specifies alternative outcome sets, a probability measure, and a utility function over world states. Plans generated should maximize the expected utility.

We propose a third technique, which we call *epsilon-safe ($\epsilon$-safe) planning*, positioned between these two extremes. The intention is as follows: the planning system will make use of conditional planning techniques, using information about the likelihoods of the various different outcomes of the conditional acts to guide the planning process and to support the effective construction of incomplete conditional plans. The planner will attempt to provide a plan which will achieve the goal with a probability at least $1 - \epsilon$, where $\epsilon$ is a parameter to be supplied by the user.[1]

While probabilistic reasoning adds complexity to the representations used in conditional planning, it is our thesis that in practice this complexity will result in a simplification of the actual planning process. The use of probabilities allows us to impose the $\epsilon$ cutoff, allowing us to ignore low-probability portions of the search space. Probabilities also provide us with an effective heuristic (an optimistic estimate of the probability of plan success) for searching the space.

While we agree that decision-theoretic planning is, in an ideal sense, the Right Thing, we feel that much work remains to be done before decision theoretic techniques may be combined with planners. Wellman and Doyle [1992] and more recently Haddawy and Hanks [1993] have shown that goals as used by AI planners cannot be straightforwardly (i.e., without human intervention) translated into utility functions.

Our approach is inspired by techniques used in engineering risky systems where it is regarded as either infeasible or undesirable to specify a utility model. Under such circumstances, one typically specifies that one

---

[1] Kushmerick, Hanks and Weld [1993] have independently developed an approach similar to epsilon-safe planning, though the resulting plans are not conditional. Draper, et. al. [1993] are extending this framework to conditional plans.

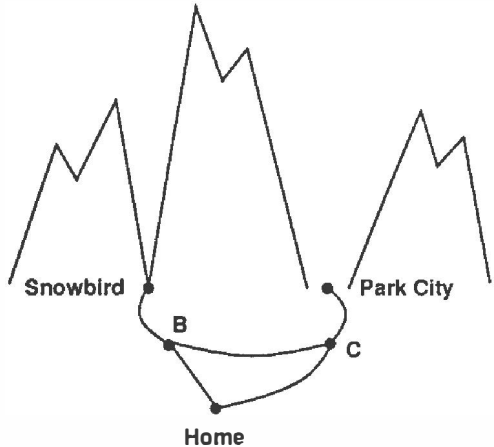

Figure 1: The Skiier's micro-world.

wants a system which has a failure bound of epsilon over some period of use.

In this paper, we present two simple $\epsilon$-safe planners constructed as extensions to two conditional planners: PLINTH [Goldman and Boddy, 1994a] and CNLP [Peot and Smith, 1992]. These conditional planners generate plans in the presence of actions with uncertain outcomes. We show that an epsilon-safe planner may be constructed atop them in a straightforward way, given that one assumes that the outcomes of conditional actions are chosen independently. We then show how this independence assumption may be relaxed. To do so, we combine techniques of conditional planning with knowledge-based construction of probabilistic models.

## 2   CNLP

Recently, Peot and Smith [1992] have introduced CNLP, a conditional planner based on McAllester and Rosenblitt's SNLP [McAllester and Rosenblitt, 1991]. CNLP extends SNLP by adding *conditional actions*. Conditional actions provide a simple, disjunctive representation of uncertainty. Each conditional action has a number of possible outcomes, one of which will occur if the action is executed.[2] Given a goal, an initial state and a set of actions (some of which may be conditional) the CNLP algorithm will generate a conditional plan to achieve the goal. The conditional plan is akin to a conventional non-linear plan, but it branches at every conditional action. Below each branch point will be a partial conditional plan which will carry the agent from the given outcome to the goal. For example, Figure 2(a), taken from Peot and Smith's paper, gives a conditional plan for going skiing in the situation depicted in Figure 1.

CNLP is an elaboration of SNLP. SNLP constructs a non-linear plan by building a directed acyclic graph of operators between distinguished start and finish nodes. Any ordering of the actions which represents a topological sort of the DAG will achieve the goal (modulo the standard classical planning assumptions). The edges of the graph are either *causal links* or *ordering links*. For every precondition of an action, there will be a causal link from the operator which establishes that precondition to the action which has it as a precondition. Ordering links are employed to eliminate potential clobberers (steps which delete protected propositions).

---
[2]We provide a detailed analysis of conditional actions elsewhere [Goldman and Boddy, 1994b].

CNLP generalizes the graph representing a (partial) nonlinear plan by imposing on it a notion of *context*. When a conditional action is introduced into the plan graph, it splits the plan into multiple contexts: one for each outcome of the conditional action. For example, the action (**observe (road b s)**) has two possible outcomes: (**clear b s**) and (**not (clear b s)**). For each context, CNLP introduces a unique label. As in Figure 2(a), let us call the label for (**clear b s**) $ol_1$ and that for (**not (clear b s)**), $ol_2$. Labels are used to mark the various actions of the plan to indicate the conditions under which they should be performed. This label is propagated to every proposition and action which depend on the conditional outcome. To each context introduced, there corresponds a separate goal node. This has the effect of creating a separate plan for every context. In the skiing example, when the conditional action (**observe (road b s)**) is introduced into the plan, a second goal node is introduced. The original goal node is now labeled to indicate that it corresponds to the case where the road is seen to be clear and a second goal node is introduced for the problem of planning a solution to the problem which will work when the road is known not to be clear.

The CNLP algorithm involves only limited extensions to SNLP. CNLP only needs to worry about threats within compatible contexts. CNLP may resolve threats not only by ordering them before or after matching preconditions, but also by forcing them into different contexts.

## 3   Plinth

In work reported elsewhere, we have developed PLINTH [Goldman and Boddy, 1994a], a linear conditional planner loosely based on McDermott's regression planner PEDESTAL [McDermott, 1991]. We have shown that this planner is sound and complete with respect to its action representation.

Given the current prevalence and popularity of nonlinear planning, our decision to construct a linear conditional planner may require some explanation. In conventional, "classical" planning applications, nonlinear planning is usually an improvement over linear planning because fewer commitments yields a smaller search space, at a small added cost to explore each element of that search space [Minton et al., 1991]. How-ever, it is not clear that this tradeoff operates in the same way for conditional planners. When plans have multiple branches, the savings from considering fewer orderings is likely to be much less and may not repay the cost in the added complexity of individual plan expansion actions. In particular, the domain in which we have applied PLINTH is one in which subgoal interactions are minor, and thus in which a linear planner can be effectively employed. Conditional linear planning is simpler in conception as well as in implementation. In particular, our conditional linear planner can be shown to be sound and complete; we do not yet know of a sound and complete conditional nonlinear planner. Finally, the operation which is needed to properly construct branching non-linear plans — resolving clobberers through conditioning apart — is a very difficult operation to direct. In addition to the arguments on behalf of a conditional linear planner in and of itself, imposing a probability structure over a conditional linear plan is easier than imposing it over a conditional nonlinear plan.

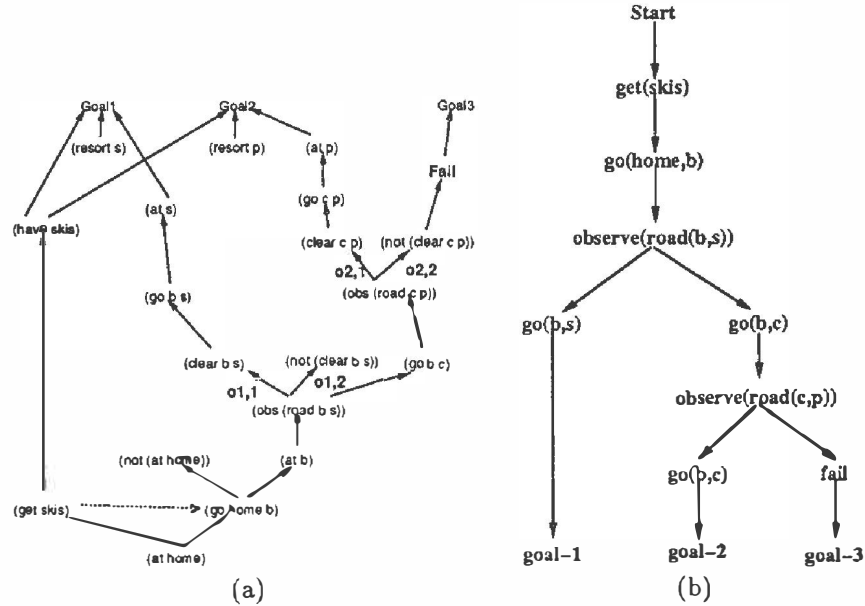

Figure 2: Conditional plans to go to a ski-resort. (a) A CNLP plan. Causal links are shown as solid lines. Multiple possible outcomes of an action are marked as o1, o2. Broken lines are ordering links. (b) A PLINTH plan.

PLINTH's conditional linear planning algorithm is non-deterministic and regressive. The planner operates by selecting an unrealized goal and nondeterministically choosing a way to resolve that goal while respecting existing protections. To resolve a goal PLINTH will either

1. find that goal to be true in the initial state;
2. find an existing step in the plan which establishes the goal or
3. add to the plan a new step which establishes the goal.

New goals may be introduced when steps are introduced, either to satisfy preconditions or to plan for contingencies introduced by conditional actions. In essence, this algorithm is the same as that of a conventional linear planner. The crucial difference is in the effect of adding a conditional action to the plan.

When adding a conditional action, $A$, there will be some outcome, $O$, such that $A - O$ will establish the goal literal (otherwise $A$ would not have been chosen for insertion). This outcome will establish what one can think of as the "main line" of the plan. However, there will also be some set of alternative outcomes, $\{O_i\}$. In order to derive a plan which is guaranteed to achieve the goal, one must find a set of actions which can be added to the plan such that the goals are achieved after $A - O_i$. for all $i$. This is done by adding new goal nodes to the tree for these alternative contingencies. The addition of these goal nodes renders the plan tree-shaped, rather than linear. PLINTH will plan for all of the goal nodes in the plan. The early portions of the plan will be shared among the different contingencies.

A PLINTH plan for the earlier Ski World example is shown in in Figure 2(b).

## 4 $\epsilon$-safe planning with simple model

A preliminary approach to probabilistic planning may be developed by imposing a simple probability measure over partial plans. We start by associating with each conditional operator a probability distribution over its outcomes. The plan tree will provide a proba-bility measure of the chance of being at a given point in the plan. We use this measure in two ways: first, by computing the probability mass associated with the set of goal nodes in the plan graph, we can determine an estimated probability of success of our plan; second, when choosing a next node to expand in our search, we will choose to expand one whose label has the most probability mass and thus can contribute the most probability mass to the successful plan. We may construct this measure in a straightforward way if we assume that the outcomes of the conditional actions are independent.

The two conditional planners admit of a simple epsilon-safe adaptation if one makes two assumptions:

1. Actions will only be done when their preconditions are known to be true;
2. The probability distribution over the outcomes of an action depends only on the state of the world encapsulated in the preconditions for that action.

The first is a commonly-made planning assumption. The second is a Markov assumption, one of the consequences of which is that conditional action outcomes are conditionally independent, given the preconditions encoded in each operator. In our construction of more complicated probability models (Section 5), both of these assumptions are relaxed somewhat.

To build such an adaptation, one must be able to assign a probability measure over a conditional plan. We start by associating with each conditional action a distribution over its outcomes. Given these distributions and the assumptions above, we can impose a probability measure on PLINTH plans. Recall that a PLINTH plan tree contains multiple goal nodes, each of which corresponds to a context — a single set of outcomes of conditional actions. Since the contexts are mutually exclusive and exhaustive, we can add together the probability of success for each completed goal node to determine the overall probability of success. Since the outcome of each conditional action is independent, we can associate with each label the corresponding outcome probability and determine the probability of success by multiplying together the probabilities associated with each outcome label in each path to a goal node. An analogous method suffices to put a probability measure on CNLP plans.

An important feature of the probability measure is that even before the plan is completed, it can be used to bound the probability of success. In particular, we can choose which goal node (context) to work on based on how promising it is. We believe that this heuristic information will substantially improve planning search.

We may work around our Markov assumption to some extent, at the expense of some effort in encoding the operators and world state. Consider a case where the independence assumption taken at its face value is clear to be violated. In Peot and Smith's SkiWorld example, the probability of observing the roads to be open could be radically changed by the occurrence of a blizzard, in a way which induces a dependency between these outcomes. We can capture this interaction by augmenting the representation of the world state, and splitting the existing operators into multiple operators with more complex preconditions.

We can capture the dependency within the limits of the simple model by adding to the world state representation the proposition **blizzard**. Since the probability of success following a road is influenced by whether or not there is a blizzard, we replace our **observe(clear(X,Y))** operator schema by two operator schemas, one for the case where there is a blizzard, and one for the case where there is not a blizzard. The BURIDAN planner [Kushmerick et al., 1993] handles context-dependency by the use of *triggers*, an equivalent mechanism.

Representational tricks of this kind will only be sufficient to cover over limited departures from the essential model. If there are too many dependencies, the notation will become too unwieldy. Furthermore, this trick will not work in the event that there are unobservable propositions which induce dependencies.

## 5  Building more complex models

We believe that the simple probabilistic model described above will prove to be useful for domains with a simple action model. However, the simple model breaks down when applied to domains where there are significant dependencies among action outcomes, particularly if these dependencies are the result of events that are not directly observable.

Consider an elaboration of Peot and Smith's "Ski world" example. Let us suppose that points B and C are within, say, 100 miles of each other. Then in severe snowstorm conditions, there will be a substantial correlation between the states of the road from B to Snowbird and the road from C to Park City. If we have the additional option of going to Switzerland to go skiing,[3] this dependency may be quite important. Upon observing the closure of the road to Snowbird, it may be better for us to just fly to Switzerland, since we know that it is likely that the road to Park City will also be blocked. Alternatively, it may be a good idea for us to make a plan in which we first listen to the radio to determine whether or not there is a blizzard. In the event of a blizzard, we fly to Switzerland, otherwise we should try to find an acceptable route to one of the two resorts.

In this section we outline linear and nonlinear versions of the $\epsilon$-safe planner that allow for dependencies. These planners build probability models in par-allel with the construction of their plans. The probability model employs techniques of Knowledge-based model construction [Wellman *et al.*, 1992], which we will discuss further below. We begin by outlining our simplifying assumptions; we discuss the formal representation of the planning problem; we briefly review knowledge-based model construction and we conclude with a presentation of the $\epsilon$-safe planning algorithms.

---

[3] Let us assume we are also quite wealthy and do not have to be at work on Monday.

### 5.1 Assumptions

1. The initial state of the world is completely known, modulo uncertainty. For every proposition $Q$, the initial state contains either $Q$, $not(Q)$ or a prior distribution, $P(Q)$. We allow the prior distribution to be factored as a directed graph (belief network). E.g., if there are three propositions A, B and C whose truth value we do not know, we may wish to represent $P(A)$, $P(B)$ and $P(C)$ in terms, say, of $P(A)$, $P(B|A)$ and $P(C|A)$.

2. The outcomes of every action are observed by the agent. Outcomes cannot be predicted, but will be observed when they occur.

3. We assume that all observation actions are infallible. For the purposes of this preliminary study, this simplifies things considerably.

4. In order for an action to be performed, all of its preconditions must be *known* to hold.

Assumptions 2 and 4 are simplifications that make the algorithm much cleaner. That these simplifications are reasonable for applications such as planning organizational behavior.[4] If one is, say, planning the operations of a trucking company, it is reasonable to assume that one will know whether or not the truck arrives at the warehouse and whether or not it is out of fuel. These assumptions are probably not reasonable for applications such as planning a series of motions or manipulations by a robot with sensors that yield only approximate information. Uncertain observations (violations of assumption 3) may be modeled as certain observations of variables which are related probabilistically to real states of the world (for our domains, this seems preferable intuitively, as well).

We extend the plan representation by adding *causal influences*. Rather than encoding the establisher and consumer of a given proposition causal influences record a situation where the truth of the given proposition may change the outcome distribution for a conditional action. Causal influences will be represented differently in the plans of the two $\epsilon$-safe planners: implicitly in the linear version, explicitly in the nonlinear.

---

[4] Or generating high-level plans for "robustification" by methods like those described by Drummond and Bresina [1990]

```
blizzard({true, false})    P(blizzard) = 0.1
clear(X,Y, {true, false})  ← blizzard({true, false})
                           P(clear(X,Y)) | blizzard) = 0.1
                           P(clear(X,Y)) | not blizzard = 0.999
```

Figure 3: *Conditional outcome statements for use in the Ski World.*

### 5.2 Knowledge-based model construction (KBMC)

There exist a number of techniques for graphically representing probability and decision models (see [Pearl, 1988, Shafer and Pearl, 1990]). While these representations make possible efficient inference and knowledge acquisition, they are inherently propositional and are limited in their abilities to represent particular problem instances. The KBMC approach is to encode general knowledge in an expressive language, then construct a decision model for each particular situation or problem instance. The interested reader can find more details on KBMC in the review by Wellman, et. al. [1992]. For the purposes of our $\epsilon$-safe planner, we record knowledge about patterns of causal interaction in a knowledge-base. When attacking any particular planning problem, relevant pieces of this knowledge will be drawn from the knowledge base and assembled into a probability model to evaluate the conditional outcomes.

Breese's ALTERID system [Breese, 1992] provides a complete and sound method for KBMC. In ALTERID, random variables are named by conditional outcome statements. Conditional probabilities of various outcomes are described using a notation similar to Horn clauses. As in Horn clauses, there is a head, in this case naming the influenced variable. The body of the clause is a list of conditional outcome statements which influence the head variable. If the body of the clause is empty an unconditional distribution must be provided. For example, we might encode our (very limited) knowledge of meteorology in the Ski domain using the ALTERID clauses given in Figure 3.

In $\epsilon$-safe planners, plans have a bipartite representation. In addition to the plan graph, there is also a probabilistic model. This is a belief network, whose start state is constructed from the initial conditions, and which will be augmented to reflect the current partial plan.

### 5.3 Linear Planning

As in PLINTH, plans will be represented as trees. In PLINTH, the nodes in the plan graph are plan *steps*, each of which is a particular instance of some operator. There are also two kinds of dummy steps, **start**, which roots the tree, and **goal**. Goal nodes have as preconditions the goal propositions; every leaf of the plan tree will be a goal node.

The $\epsilon$-safe linear planning algorithm is an extension to the PLINTH algorithm. As before, goals will be introduced and discharged in one of three ways. The difference is that the algorithm will now maintain a set of causal influences to be resolved as well as a set of goals. Causal influences need to be handled specially because they may be known to be true, known to be false, or unknown, at the time the conditional action is performed. This distinguishes them from conventional preconditions, which must be known to be true.[5]

There are three ways to discharge an open causal influence:

1. Try to get to know the outcome of the event. Choose some observation action which can establish the value of the event. In the SkiWorld example, one might listen to the radio to determine whether or not there is a blizzard in the mountains.

2. Cause the event to have a particular outcome. A somewhat farfetched example would be to seed the clouds to ensure the presence of a blizzard. Etzioni, et. al. [1992] discuss the issues encountered when planning in domains where propositions may be either observed or established.

    In fact, the STRIPS-style representation we use is not sufficiently expressive to capture the distinction between observation and establishment, so these two cases collapse into one from the standpoint of the planning algorithm.

3. Choose to act in ignorance of the outcome. To preserve correctness, the agent's ignorance must be protected. This is done to avoid paradoxical situations where the agent will observe a proposition, not like the outcome, so "pretend" that it hasn't made the observation and look repeatedly until the outcome is more to its liking.

Our planning algorithm, so far, is a straightforward extension of PLINTH. However, only half of our job is done. In order for this method to be useful, we need some way of calculating the probability of success of various plan branches. We use the belief net model we construct, in conjunction with the plan, to do so.

The interface between the plan- and model-building aspects of the planner will be mediated by the addition of conditional and observation actions to the plan. In doing so, we will construct a probability distribution over the contexts of the plan.

On invocation of the planner, we will use the ALTERID algorithm and the initial state description to construct a belief network capturing the prior distribution over the propositions whose truth values are unknown.[6] Each outcome of a conditional outcome statement will have a unique label. Labels in the belief network will correspond to labels in the plan.

When conditional steps are added to the plan, parallel nodes will be added to the belief network. The outcome space of this node is the set of possible outcomes of the conditional step. For every *open* influence on the conditional action, we draw an arc from the corresponding conditional outcome statement in the belief network to the conditional action node. I.e., the arcs in the belief network represent those causal influences which must be encapsulated in a probability distribution.

No nodes need be added to the belief network to parallel the addition of observation steps to the plan. However, the labels of the observation outcomes must be the labels of the outcomes of the observed variable in the belief network.

As causal influences are resolved these causal links will be cut. Now, in order to find the probability of a context, one finds the joint probability of a set of labels (which are outcomes of variables in the network). Since the goal nodes are labeled with contexts, one may use this computation to determine an upper bound on the probability of success of a plan branch. To find the probability of a particular outcome in a plan branch, find the probability of the outcome conditional on the probability of the context in which the action is executed.

Let us return to the Ski World example,[7] complicated by the dependencies given in Figure 3 and consider one way the algorithm might proceed. Initially, the planner attempts to reach Snowbird, reaching the state shown in Figure 4(a). This is done by first discharging the goals to reach Snowbird, then the subgoal of the road from B to Snowbird being clear. When this latter goal is discharged by adding the **observe(B,S)** step, the **clear(B,S)** node is added to the model, and we add the influence of **blizzard**. In this example, the planner has decided *not* to try to determine whether or not there is a blizzard (or perhaps is unable to do so). A protection of the form **unk(blizzard,start,observe(B,S))** must be added to the plan. The planner may evaluate the model to estimate (upper-bound) the probability of success in reaching **End1**, which is .9091.

Let us assume now that the planner successfully completes the full plan up to **End1** (given the earlier probability estimate, this is surely a sensible strategy). Now it begins to try to reach **End2** by getting to Park City. Discharging the goal of being at Park City, it adds the step of **go(C, Park City)**, which introduces the new goals of being at C and having the road from C to Park City be clear. The planner discharges the latter goal by introducing the step of observing the road from C to Park City. In turn, this introduces the influence of blizzard on clear(C, Park City). After connecting those variables in the model, we reach the situation shown in Figure 4(b). At this point, we may consult the model to bound the chance of success pursuing **End2** (which is the joint probability of finding the road from B to S closed but that from C to P open). The incremental improvement pursuing this plan is less than 1%, so if the planner is sufficiently fanatic in its devotion to skiing, perhaps it should begin considering that trip to Switzerland.

---

[5]Even ADL's secondary preconditions must have a known value at the time an action is performed.

[6]In the implementation, we will construct the belief network on a lazy, as-needed basis.

[7]In the interest of simplicity, we will ignore the have(skis) subgoal in this example.

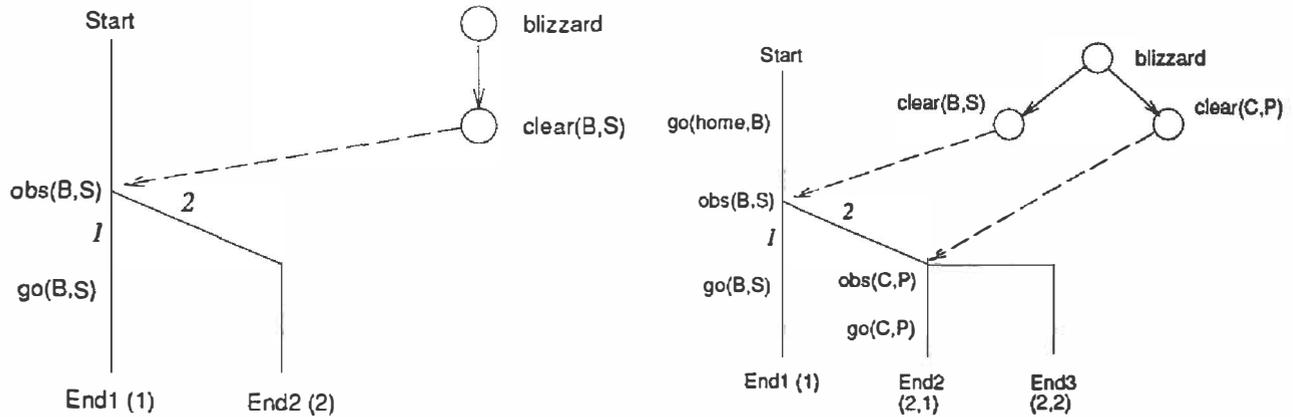

Figure 4: Example of Epsilon-planning with PLINTH.

### 5.4 Nonlinear Planning

As in CNLP, plans will be represented as directed acyclic graphs. In CNLP, the nodes in the plan graph are plan *steps*, each of which is a particular instance of some operator. There are also two dummy steps, start, which establishes the initial conditions, and finish, which has as preconditions the goal propositions. There will also be four kinds of link in the plan:

1. causal links: a triple $(e, P, c)$ (sometimes written $e \xrightarrow{P} c$), where $e$ is a step which *establishes* (has as postcondition) the proposition $P$, which is *consumed* by (appears as precondition of) the step $c$.

2. conditioning links: a triple $\{C, \alpha, B\}$, where $C$ is a conditional action, one of whose outcomes is $\alpha$ and $B$ is an action (of any type), which cannot be performed unless the outcome of $C$ is $\alpha$.

3. ordering constraints: A pair $(S_1, S_2)$ (commonly written $S_1 < S_2$) which indicates that step $S_1$ must be performed before $S_2$.

4. influence links: A triple $[i, P, c]$, where $i$ is a step and $P$ is in a proposition family which influences the conditional step $c$.

The first three types of links are as per CNLP. The latter is new to the $\epsilon$-safe planner.

In order to reflect the fact that the planner's uncertainty about the world is encoded in the initial conditions, we treat ignorance preconditions specially. We have a special kind of link: any step, $s$ which has as its precondition that some the outcome of some conditional outcome statement, $P$, be unknown, may have that precondition satisfied only by a link of the form (start, $P, s$. This is a consequence of the fact that ignorance is never *added* to an $\epsilon$-safe planning problem (a consequence of our simplifying assumptions).

The $\epsilon$-safe planning algorithm is given in Appendix A. Note that most of the algorithm is identical to the CNLP algorithm. We have simply added a new step, necessary to properly handle causal influences on conditional actions. Causal influences need to be handled specially because they may be known to be true, known to be false, or unknown, at the time the conditional action is performed. This distinguishes them from conventional preconditions, which must be known to be true.[8]

This added complication is handled in step 4 of the Find-completion procedure. One way to handle an open causal influence is to try to get to know the outcome of the event. Step 4a handles collecting information about the outcome.

The somewhat unusual aspect is provided by step 4b. The purpose of this step is to allow us to handle the case where the planner intends to perform a conditional action in ignorance of the outcome of some causal influences. To preserve correctness, the agent's ignorance must be protected. This is done to avoid paradoxical situations where the agent will observe a proposition, not like the outcome, so "pretend" that it hasn't made the observation. For example, let us assume the status of a road from B to Snowbird depends on whether there has been a blizzard and the agent plans to listen to a weather report and drive down the road if no blizzard is reported. It does not make sense for the agent to plan to listen to *another* weather report, hoping for no blizzard, if it hears an unfavorable report initially. The "ignorance link" constructed in step 4b prevents this from happening.

---
[8] Even ADL's secondary preconditions must have a known value at the time an action is performed.

The interface between the model- and plan-construction operations will be handled in the same way as in the linear $\epsilon$-safe planner.

## 6   Summary

We have presented an approach to high-level planning under uncertainty that we call $\epsilon$-*safe planning*. This approach ducks the complexities of decision-theoretic planning by specifying a level of acceptable risk, $\epsilon$. The planner is committed to meet some specified goal with a probability of success of at least $1 - \epsilon$. We have presented two measures of probability for $\epsilon$-safe planning: a straightforward extension to conditional planning for which computing the necessary probabilities is simple but which makes a drastic independence assumption and a second approach which relaxes this independence assumption. The latter approach involves the incremental construction of a probability dependence model in conjunction with the construction of the plan graph. We have shown how these probability models may be integrated into PLINTH and CNLP. We are currently working on the implementation of these techniques as part of a project on planning image processing actions for NASA's Earth Observing System in collaboration with Nick Short, Jr. and Jacqueline LeMoigne-Stewart of NASA Goddard [Boddy *et al.*, 1994].

## A   $\epsilon$-safe planning algorithm for CNLP

Plan (InitialConds, Goal)

1. model := belief-net(InitialConds);
   This step of the planning algorithm establishes a model which will satisfy the constraints on the initial conditions (Assumption 1).
2. initial-plan := make-init-plan(Goal);
   We initialize the conditional plan graph.
3. Complete(initial-plan, model);
   Recursive function which finds and returns the plan (or signals failure).

Complete(Plan, Model)

1. Check for success/failure
2. If there is a link $l = (s, P, w) \cup [s, P, w]$ and there is a compatible threat $v^9$ to $l$ such that the Plan does not contain either $v > w$ or $s > v$ resolve the clobberer by doing one of the following:
   (a) **return** Complete(Plan $\cup\, v > w$)
   (b) **return** Complete(Plan $\cup\, v < s$)
   (c) Conditioning:

      i. select some conditional step $a$ which is possibly before both $v$ and $s$.
      ii. Let $c_a$ be the context of $a$.
      iii. Select two of the outcome labels of $a$, $\alpha_i$ and $\alpha_j$ ($i \neq j$), such that $c_a \cup \alpha_i$ is consistent with the context of $v$ and $c_a \cup \alpha_j$ is consistent with the context of $s$.
      iv. **return**   Complete(Plan $\cup\, \{a, \alpha_i, v\} \cup \{a, \alpha_j, s\}$);

3. If there is some step $w$ in the Plan which has some open precondition $P$. Do one of the following:

   (a) Let $s$ be some step consistent with $w$ which establishes $P$.
      **return** Complete(Plan $\cup (s, P, w)$)
   (b) Select an operator $o_i$ from the set of operators which add $P$. Create a new step $s$ which is of type $o_i$.
      **return** Complete(Plan $\cup\, s \cup (s, P, w)$)

4. If there is an open influence $P$ on a conditional step $s$, non-deterministically choose to do one of the following:

   (a) **Observe the state of the conditional outcome statement**, $P$: This may be done either using a pre-existing step of the plan, or by inserting a new step. Non-deterministically choose one of:

      i. Let $s$ be some step consistent with $w$ which establishes/observes some outcome $P_i$ in the family of $P$.
         **return** Complete(Plan $\cup (s, P_i, w)$)
      ii. Select an operator $o_i$ from the set of operators which establish/observe $P_i$, for some $P_i$ in the family of $P$. Create a new step $s$ which is of type $o_i$.
         **return** Complete(Plan $\cup\, s \cup (s, P_i, w)$)

   (b) **Record ignorance of the state of the conditional outcome statement**, $P$: If the outcome of $P$ is unknown in the initial state,
      **return**
      Complete(Plan $\cup$ [start, Unknown($P$), $w$])

5. **Collect additional information:** For some conditional step, $s$, for some influence $P$ on $s$ such that $P$ is unknown at the time $s$ will be done, choose some operator $o$, (one of) whose outcome(s) is $P'$ which is not independent of $P$. Create step $s'$, instance of $o$,
   **return** Complete(Plan $\cup\, s' \cup (s', P', s)$)

6. If there is a post condition with a context $c_{\text{new}}$ which is not compatible with the context of any existing goal, create a new goal step $\mathcal{G}$ whose context is $c_{\text{new}}$.
   **return** Complete(Plan $\cup\, \mathcal{G}$)

---

[9]Note that if $P$ is of the form Unknown($X$), then $P$ will be threatened by any $X_i$ in the family of $X$.

## Acknowledgements

Thanks to Mark Peot, and David Smith for discussions which helped clarify these ideas, and to the anonymous reviews for helpful comments.